%% file: sample-sigconf.tex
\pdfoutput=1 %This line was added to correct arxiv processing error

\documentclass[sigconf,final]{acmart}

\usepackage{booktabs} % For formal tables
\usepackage{url}

\usepackage{graphicx}
\usepackage{subcaption}
\usepackage{csquotes}

\setcopyright{acmlicensed} 

\begin{document}
\title{Leveling the Playing Field}
\subtitle{Fairness in AI Versus Human Game Benchmarks}
% \titlenote{Produces the permission block, and
% \subtitle{Exploring Patterns in 2D Tile-based Games}
% \subtitlenote{The full version of the author's guide is available as
%   \texttt{acmart.pdf} document}

% \author{Anonymous}
% \affiliation{%
%   \institution{Some Place}
%   \streetaddress{Street}
%   \city{City}
%   \state{State}
%   \postcode{12345}
% }
% \email{email@email.com}

\author{Rodrigo Canaan}
\affiliation{%
  \institution{Department of Computer Science and
Engineering,\\ New York University}
  \streetaddress{5 Metrotech Center}
  \city{Brooklyn}
  \state{NY}
  \postcode{11201}
}
\email{rodrigo.canaan@nyu.edu}

\author{Christoph Salge}
\orcid{1234-5678-9012}
\affiliation{%
  \institution{School of Computer Science,\\ University of Hertfordshire}
  \streetaddress{College Lane}
  \city{Hatfield}
  \state{UK}
  \postcode{AL10 9AB}
}
% \affiliation{%
%   \institution{Department of Computer Science and
% Engineering,\\ New York University}
%   \streetaddress{5 Metrotech Center}
%   \city{Brooklyn}
%   \state{NY}
%   \postcode{11201}
% }
\email{ChristophSalge@gmail.com}

\author{Julian Togelius}
\affiliation{%
  \institution{Department of Computer Science and
Engineering,\\ New York University}
  \streetaddress{5 Metrotech Center}
  \city{Brooklyn}
  \state{NY}
  \postcode{11201}
}
\email{julian@togelius.com}

\author{Andy Nealen}
\affiliation{%
  \institution{School of Cinematic Arts,\\ University of Southern California}
  \streetaddress{900 West 34th Street}
  \city{Los Angeles}
  \state{CA}
  \postcode{90089-2211}
}
\email{anealen@cinema.usc.edu}

\renewcommand{\shortauthors}{Canaan et al.}

\begin{abstract}
From the  beginning of the history of AI, there has been interest in games as a platform of research. As the field developed, human-level competence in complex games became a target researchers worked to reach. Only relatively recently has this target been finally met for traditional tabletop games such as Backgammon, Chess and Go. This prompted a shift in research focus towards electronic games, which provide unique new challenges. As is often the case with AI research, these results are liable to be exaggerated or misrepresented by either authors or third parties. The extent to which these game benchmarks constitute \enquote{fair} competition between human and AI is also a matter of debate. In this paper, we review statements made by reseachers and third parties in the general media and academic publications about these game benchmark results. We analyze what a fair competition would look like and suggest a taxonomy of dimensions to frame the debate of fairness in game contests between humans and machines. Eventually, we argue that there is no completely fair way to compare human and AI performance on a game.
\end{abstract}




\copyrightyear{2019} 
\acmYear{2019} 
\acmConference[FDG '19]{The Fourteenth International Conference on the Foundations of Digital Games}{August 26--30, 2019}{San Luis Obispo, CA, USA}
\acmBooktitle{The Fourteenth International Conference on the Foundations of Digital Games (FDG '19), August 26--30, 2019, San Luis Obispo, CA, USA}
\acmPrice{15.00}
\acmDOI{10.1145/3337722.3337750}
\acmISBN{978-1-4503-7217-6/19/08}

\maketitle
\input{samplebody-conf}

\bibliographystyle{ACM-Reference-Format}
\bibliography{sample-bibliography}

\end{document}

%% file: samplebody-conf.tex
\pdfoutput=1 %This line was added to correct arxiv processing error

\section{INTRODUCTION}

Games have a long history of being used as Artificial Intelligence testbeds and benchmarks. The formalized representation, especially of board games, and the often explicit and clear reward structure makes them well suited for AI approaches. With recent advances in AI research there is also an emergent narrative that general game playing, i.e. playing a range of different games with the same agent, is a necessary stepping stone towards Artificial General Intelligence (AGI)~\cite{yannakakis2018artificial,schaul2011measuring}. Setting this question aside, we want to focus on another related question: Are games a good way to test if an AI has human-level intelligence?

This question is often brought up, given that humans are widely accepted to be the only known example of general intelligence. Surpassing or at least achieving parity with human-level artificial intelligence seems to be a necessary step to reach AGI, and what better way to demonstrate this then to beat a human (or the best human) in a \enquote{fair} competition of intelligence, i.e. beat them in a game? Consequently, the media, and to some extent the scientific literature, often characterizes human-AI game competition as a way to determine if an AI has finally reached human-level intelligence.

The central argument presented in this paper objects to this characterization. We argue that there is no, and that there can never be a \enquote{fair} comparison between an AI and a human that answers the question if an AI has human-level intelligence. Building up to this argument, we will first give a short survey of how human-AI game competitions are currently portrayed, both in the media and in scientific literature.  We will then look at several existing examples of human-AI game comparisons and demonstrate why it is hard to establish a fair comparison for those specific examples. Based on these examples we will then make a more general argument, outlining how the range of different AIs is just too wide to find any game that offers a fair comparison. As a corollary, we can say that for something to have human-level AI in a meaningful way it would have to be human, or very close to human, in both physical, mental and social embedding.

\section{Portrayal of AI game benchmark achievements}
\label{sec:portrayal}

To set the stage for our discussion about benchmarks comparing AI and humans, we will look at important AI benchmarks in classic board games such as Backgammon, Chess an Go, as well as modern electronic games. We will inspect the claims that were made by the original authors of the systems that achieved success in these game benchmarks, and how those results were subsequently discussed in the general media and follow-up academic papers. As an analysis tool, we will use previously published guidelines on how to write about AI research such as ~\cite{SevenSins} and~\cite{Unhinged}. Our goal in this section is to illustrate how game AI benchmarks are perceived by society, and what the main concerns regarding the fairness of comparison between human and AI programs are.

%While the use of \enquote{suitcase words}~\cite{SevenSins} is almost unavoidable given the prevalence of terms such as intelligence, learning, prediction even academic writing, more egregious violations will be explicitly noted. 

Note that the examples gathered in this sections are not meant as a statistically representative sample of all that has been written about each of these achievements.
\subsection{TD Gammon}
\label{sec:TD Gammon}

TD-Gammon~\cite{tesauro1995temporal} is a Backgammon-playing program developed by Gerald Tesauro at IBM using temporal-difference learning, a reinforcement learning technique where a neural network is trained through self-play to minimize the difference between its prediction of the game's outcome and the actual outcome over successive game states. Between 1991 and 1992, it played over a hundred games against some of the best players in the world across three different versions of the software. The last version (TGD 2.1) came very close to parity with Bill Robertie, a former world champion, losing a 40-game series by a difference of a single point.

Tesauro highlights that observing the algorithm play has led to a change in how humans evaluate positions, especially in opening theory for the game. In particular, with some opening rolls, the system preferred "splitting" its back checkers rather than the more risky, but favored at the time, option of "slotting" its 5-point. Since then, the splitting opening has been confirmed as the superior choice by computer rollouts and is now the standard choice for the 2-1, 4-1 and 5-1 initial rolls.

When discussing  applicability in other domains, Tesauro lists robot motor control and financial trading as potential applications while cautioning that the lack of a forward model and the scarcity of data might limit the success in these real world environments. Not much discussion of TD-Gammon's achievements was found in general media dating from the time of its release, but Woolsey, an analyst in Tesauro's paper~\cite{tesauro1995temporal} states that TD-Gammon's algorithm is \enquote{smart} and learns \enquote{pretty much the same way humans do}, as opposed to \enquote{dumb} chess programs that merely calculate faster than humans. It is interesting from a historical perspective to see that the distinction between real or human-like intelligence and mere calculations was already a concern even before Deep Blue's success in Chess, which we examine next, brought the issue to the forefront.

\subsection{Deep Blue}
\label{sec:Deep Blue}

Deep Blue~\cite{campbell2002deep} is a program and purpose-built computer for playing chess, designed by a team at IBM led by Murray Campbell. It uses a combination of specialized hardware and software, such as tree-search augmented by heuristics for pruning and state evaluation crafted by human experts. It also uses a database of opening moves and endgame scenarios to select moves at the start and end of a match with little computational effort. It achieved enormous visibility in 1997 when it defeated the reigning champion Garry Kasparov in a six-game match with tournament rules by a score of $3\frac{1}{2} - 2\frac{1}{2}$. Kasparov had previously beaten a former version of the algorithm in 1996 by a score of $4 -2$.

In their paper describing the system~\cite{campbell2002deep}, the authors refrain from making speculative claims about the algorithm or its impact on the future of AI.  However, the same cannot be said about the media. One article from the Weekly Standard, with the ominous title \enquote{Be Afraid}~\cite{BeAfraid}, on the one hand claims that the system's \enquote{brute force} approach in the first game is mere calculation and \enquote{not artificial intelligence}, but on the other hand attributes brilliance, creativity and humanity to a win in the second game from a position that allegedly did not benefit as much from brute-force calculation. They also speculate on the real-world applications of Artificial Intelligence by imagining a scenario where machines might become \enquote{creatures sharing our planet who not only imitate and surpass us in logic, who have perhaps even achieved consciousness and free will, but are utterly devoid of the kind of feelings and emotions that, literally, humanize human beings}.  %This claim is backed by an interview of Deep Blue programmer Joe Hoane in the same article.% 

% From there, however, the article argues that in the second game, \enquote{Deep Blue won. Brilliantly. Creatively. Humanly} from a position that allegedly does not benefit as much from brute-force calculation.
% Then, they speculate that this amounts to passing  a chess-specific Turing test and, if machines can pass this test, they might eventually pass the more general Turing test and grow beyond our control and understanding. Ultimately, they might become 

% This is an example of magical thinking and \enquote{Hollywood scenario}~\cite{SevenSins}, where AI might gain human-like abilities in general intelligence through unspecified means unrelated to Chess research, and whose potentially dangerous results should be feared.
% and in this view, the competition is \enquote{not a matter of man versus machine but machine versus machine}.

Other commentators, such as in this New York Times article~\cite{Deepest} characterize both Deep Blue and the human brain as information processing machines, with the main difference being that Kasparov and humans have feelings such as fear and regret. While Deep Blue has no such feelings, the authors speculate that in the future, more sophisticated machines (which they name \enquote{Deeper Blue} and \enquote{Deepest Blue}) might be able to model its opponent and even have life goals outside of chess, such becoming famous, and even achieve consciousness. Ironically, they speculate that the machine might then be vulnerable to psychological warfare, at which point humans would again stand a chance in a game of chess.

Another article, also from the New York Times~\cite{Swift} highlights Kasparov's own comments not only on the psychological differences such as fear, but also physical differences such as the need to rest. Kasparov also argues that previous games by Deep Blue should have been made available prior to the match. This final remark might be justified by the fact that Deep Blue's \enquote{opening preparation was most extensive in those openings expected to arise in match play against Kasparov}, although ultimately \enquote{none of the Kasparov-specific preparation arose in the 1997 match.}~\cite{campbell2002deep}.

The biggest takeaway from Deep Blue's success is that the game of chess had held, to some extent, the status of a \enquote{representative measure of both human and computer intelligence}~\cite{ensmenger2012chess}. When a conceptually simple algorithm was finally able to convincingly beat one of the best players of its time, the implications of this feat were widely discussed, including whether or not the victory was achieved under fair circumstances, considering the differences among the systems (human feelings, exhaustion), the amount of preparation time and the nature of the underlying reasoning systems, that is, the difference between supposedly sophisticated human reasoning versus brute-force calculations.

% From these two articles, we see how the success of AI chess (which, for humans, requires intelligence) quickly invites speculation about other features of human condition, such as creativity, emotions and consciousness, even though the underlying process of the AI is viewed by some as \enquote{mere calculation}. We also see how super-human performance in one task automatically invites the thought that one day AI might achieve super-human performance in all tasks, which in some cases is painted as a scenario to be feared.

% % Another article, ffocus on Kasparov's own reactions to the match, especially the last one, which Kasparov conceded after 19 moves claiming he had lost his fighting spirit and that he, as a human being, is afraid when faced with something he does not understand. Kasparov also said that the match should have been longer, as he needs time to rest, and that previous games by Deep Blue should be made available. 
%  In this article, we see a more nuanced discussion about whether the competition between Deep Blue and Kasparov was fair, in terms of psychology, fatigue and the availability of information. These are topics to which we will come back to in our own discussion about fairness.

\subsection{Alpha Go, Alpha Go Zero and Alpha Zero}
\label{sec:Alpha Go}

In 2016, AlphaGo~\cite{silver2016mastering}, an agent developed by group of Google DeepMind researchers led by David Silver, became the first program to beat a human champion of Go in a match against Lee Sedol, in which AlphaGo won by $4 - 1$. The game had been considered one of the next big challenges for game AI after chess, due to its large search space and the difficulty to craft a good state evaluation function that accurately predicts the winner of each position.

The system achieved success through a combination of Monte Carlo Tree Search with convolutional neural networks, which learned from professional human games and self-play. In 2017, they announced a new version, AlphaGo Zero~\cite{silver2017mastering}, which learned entirely from self-play, with no human examples, and which was able to beat the previous AlphaGo version (AlphaGo Lee). Still in 2017, they announced AlphaZero~\cite{silver2017mastering2}, which uses a similar architecture (but different input representations and training) to beat other top engines in Go, Chess and Shogi.

The authors claim that the later versions of the system, (i.e., AlphaGo Zero and Alpha Zero) master the games without human help, or \textit{Tabula Rasa}. These claims were scrutinized in a paper by Gary Marcus~\cite{marcus2018innateness}, who views the agent not as learning from a \enquote{blank slate}, but as an example of hybrid system that benefits both from learned behaviors and from prior human knowledge that contributed to the system. In particular, he points out the inability of the system to generalize to variations of the game without further training. The system is also unable to learn the paradigm of tree search or the rules of the game, which humans are capable of.

In the general public sphere, AlphaGo and its successors also received wide media coverage. While Deep Blue raised questions about the nature of intelligence due to its conceptually simple \enquote{brute force} approach, discussions about AlphaGo and its successors are more focused on:
\begin{itemize}
    \item the claim that it uses comparatively less human supervision and expertise and
    \item the fact that it can learn from large amounts of data.
\end{itemize} 

The first point leads into a perception of it being both more general and less controllable, which raises questions of automation, job loss, general intelligence, creativity and the need for supervision and interpretability. The second  leads to concerns about privacy and bias when systems using similar techniques are applied to more sensitive domains.

An article from Wired~\cite{TwoMoves} acknowledges the concerns that the underlying technology behind AlphaGo could be hard to control and lead to job losses, but ultimately delivers a vision of a relationship between human and machine where they complement each other, pointing that there are still things machines cannot do, that players are now able to view the game in a different light thanks to AlphaGo and highlighting that both Lee Sedol and AlphaGo (and thus humans and machines) are able to generate \enquote{transcendent} moments, such as the now famous move 37 (by AlphaGo) and move 78 (by Sedol), both of which were evaluated by AlphaGo as having a probability of being played by a human close to one in ten million. An article by The Washington Post~\cite{Post} also looks at move 37, and asks experts about its implications for creativity. One interviewee, Pedro Domingos, sees the move as creative. Others, such as Jerry Kaplan, attribute the move to clever programming, not creativity of the software.

Another article, titled \enquote{Why is Elon Musk afraid of AlphaGo-Zero?}~\cite{WhyIsElon} takes a bleaker view, describing the advancements from AlphaGo to AlphaGo Zero as an example of the AI becoming \enquote{smart and self-aware} and creating \enquote{its own AI which was as smart as itself if not smarter}. The article goes on to wonder about the risks of such an AI being able to generalize from data in other domains, such as in the defense and military industry, and potentially becoming \enquote{unsafe}, again showing concern for what it would mean for a super-human, uncontrollable AI when applied to non-game applications, echoing the fear of the \enquote{technological singularity} hypothesized by Vinge~\cite{vinge1993technological}.

\subsection{Electronic games}
\label{sec:ALE}

Electronic games (or video games) offer additional challenges to AI researchers compared to traditional tabletop games. Due to a combination of almost continuous time scale (limited by the system's frame rate) and potentially huge game state space and action space, electronic games are typically even more intractable by brute-force search than games such as Go or Chess. As an example, an estimate by Onta\~non et al,~\cite{ontanon2013survey} estimates the state space of Starcraft as $10^{1685}$, its branching factor as $10^{50}$ and its depth as $36000$, whereas Go has corresponding values of roughly $10^{170}$, $300$ and $200$. As such, a number of video game AI benchmarks have been proposed. The use of video games as AI benchmarks goes back a long way but interest in these benchmarks has spiked since AlphaGo's results in 2016, because Go, which was considered among the most challenging tabletop games, was finally beaten and new, harder challenges had to be explored.

Some of these benchmarks encourage the development of general techniques, that can be applied for a large number of domain problems, such as different games. That is the case of  frameworks such as the Arcade Learning Environment (ALE)~\cite{bellemare2013arcade}, where agents can be evaluated in one of hundreds of Atari 2600 games and the General Video Game AI Competition~\cite{perez2016general}, where agents are evaluated in previously unseen arcade-like games.

Other examples of benchmarks proposed for specific games are Vizdoom~\cite{kempka2016vizdoom} (first person shooter), the Mario AI Benchmark~\cite{karakovskiy2012mario} (platform game) and even benchmarks not focused on winning a game, but on tasks such as designing game levels in a platform game~\cite{shaker20112010}, building adaptive settlements in Minecraft~\cite{salge2018generative} or, inspired by the Turing test, playing in a way that is indiscernible from humans~\cite{hingston2009turing}.

While all these benchmarks have garnered academic interest, none has arguably received as much general media coverage and player attention as AI challenges using Starcraft (both the Broodwar~\cite{game:Broodwar} and Starcraft II~\cite{game:SC2} versions) and Dota2~\cite{game:Dota2}.  The fact that both Starcraft and Dota 2 are popular eSports seems to have helped garner a lot of attention from the community of players.

Real-Time Strategy (RTS) games, of which Starcraft is an example, have been proposed as an AI research environment at least since 2003~\cite{buro2003real} and games of the genre have seen organized AI competitions since 2006~\cite{ontanon2013survey}. Starcraft itself has had yearly AI competitions since 2009, and gained additional popularity as a research platform  after Google DeepMind and Blizzard, the game's publisher, jointly released a reinforcement learning environment for the game in 2017~\cite{vinyals2017starcraft}. At the time of the first draft of this paper,  the state of the art had been steadily advancing, but humans still had a clear advantage in the game, even if some observers were betting on an AI victory in the near future~\cite{HumansBetterAtStarcraft}.

In the time between this paper's first draft and final version, however, DeepMind unveiled AlphaStar~\cite{alphastar}, their Reinforcement Learning agent for Starcraft 2 that managed to beat professional human players TLO and MaNa behind closed doors, before eventually losing a live showmatch to MaNa. Interestingly, between the closed door matches that were won and the live event, the developers made some changes to the agent, which previously could observe the whole map at once (but not areas covered by \enquote{fog of war}) and later were required to actively control which regions of the map were observed by moving the camera similar to how a human would. This can be seen as an effort to minimize one potential source of unfairness in the comparison with humans, by making the input and output spaces more similar to the ones experienced by human players.

The main criticism of the agent came from the fact that it was too efficient at controlling multiple units at once (also called \enquote{micro}), however Huang~\cite{Huang} argues that, because research in AlphaStar is meant to help develop techniques that help solve complex real-world problems, it is a good thing that the agent is able to find a strategy that leverages its advantages to such extent.

Agents for playing Dota 2 have also been target of recent research and seen significant developments. In 2017, an agent by OpenAI defeated Dendi, one of the best players in the world, in a limited 1v1 version of the game in a showmatch during an official Valve tournament~\cite{OpenAI1v1}. Then, a version capable of 5v5 team play defeated a team of 5 semi-professional players in the 99th percentile of skill in another showmatch in 2018~\cite{OpenAI5,OpenAIBenchmark} before eventually losing to professional players in a showmatch during The International 8~\cite{OpenAIInternational}, the biggest Dota 2 event of 2018.

As happened with Starcraft, the state-of-the art moved between this paper's first draft and final print. In April 2019, a new version of the agent was finally able to defeat OG, the world champion team, in a live showmatch~\cite{OpenAIFive}. The new version was also able to play alongside (not just against) humans. This version of the bot was made available for play with and against players around the world for 4 days, after which the agent held a win rate of 99.4\% in competitive mode, although one team of humans was able to win 10 games in a row against the bots~\cite{OpenAIFiveArena}.

A major point of debate has been the way the OpenAI agent visualizes and interacts with the game. OpenAI describes the interface in~\cite{OpenAI5} and it includes high-level features which allow the agent to "see", at any point in time, information such as the remaining health and attack value of all units in its sight. A human would have to click on each unit, one by one, to view this information. Agents can also specify its actions at a high level by selecting ability, target, offset and even a time delay (from one to four frames). A human would have to make a combination of key presses and imprecise mouse movements to achieve the same effect.

An article on Motherboard~\cite{Motherboard} has described the advantages provided to the AI as \enquote{basically cheating}, summing up that, while humans have previously been disqualified from tournaments due to the use of a mouse with illegal programmable macro actions, \enquote{Open AI Five plays like an entire team with programmable mice and telepathy}. The article also proposes that the agent should learn directly from visuals.

In a blog post~\cite{Cook}, AI researcher Mike Cook, while ultimately having a positive view on the benchmark, also comments on the interface advantages, drawing attention to some highlights of the games where, even though the agents have a reaction speed of 200ms (in theory comparable to humans), they executed key actions such as interrupting a spell or coordinating powerful abilities in a way that is seemingly impossible for humans. Cook also warned about the potential of the AI to fall prey to situations it has never encountered in its self-play (such as the technique of \enquote{creep pulling} or unusual hero lineups) and that good performance in a few facets of the game (such as teamfighting) might give the illusion of greater overall competence in the game.

A final critique against OpenAI's agents came from the number of simplifications that had to be made to tackle a game as complex as Dota 2, such as playing with a reduced hero pool, the innability to fight Roshan (a powerful NPC that typically takes risky team-wide efforts to kill, but drops a valuable reward and is often the focus of game-deciding fights between teams) and the choice to have individual invulnerable couriers per player (as opposed to a vulnerable courier shared by the entire team). These demands can be seen in game forums such as ~\cite{Liquid, Reddit} and ultimately led to OpenAI's decision to drop most restrictions (other than the limited hero pool) in preparation for the final matches at The International 8 (which OpenAI lost) and~\cite{OpenAIInternational} and against OG (which OpenAI won)~\cite{OpenAIFive}.

\section{Dimensions of Fairness}
\label{sec:taxonomy}

Fairness is a slippery concept to define. Much has been said on the subject in contexts such as political philosophy~\cite{rawls2001justice}, economic theory~\cite{kahneman1986fairness} and even social behavior between primates~\cite{brosnan2003monkeys}. A general definition of fairness is outside the scope of this paper, but, in the realm of sports and gaming, a common  stance seems to be that a competition is unfair if one side is substantially favored by circumstances that should not affect the game. 

A version of chess where players started with different pieces based on how much weight they can lift or how much money they choose to pay before the match would be considered unfair, as physical strength and monetary wealth should not affect chess. However, no one would argue that weight lifting is unfair because it gives an advantage to the strongest lifters, and some online games (not without controversy) allow players to pay real money for game content that can be used against other players. These examples show both that what exactly constitutes \enquote{fair game} is highly game-specific, and that it is often a matter of debate even within a single game community.

Peter Stone \textit{et al.}~\cite{PHILOSOC10} refer to \enquote{preserving the essence of the game} when discussing what the rules should be for a fair soccer competition between human and robot players. Preserving the characteristics that define  the essence of the game not only allows, in their view, for a competition to be seen as fair, but also makes it so that, if the robots should win, most people would agree that the robots beat human at \textit{soccer} (and not some other game that simply resembles soccer on the surface).

However, there are multiple ways in which a competition could be seen as breaking the essence of the game and allowing factors that should not affect the game to interfere with the results of the match. While Stone \textit{et al.} list the rules and restrictions they believe should apply to their competition (e.g. a robot should not be able to run much faster than a human can, and should only communicate with each other through sounds that would be perceptible to the human ear), our aim in this section is to provide a (non-exhaustive) taxonomy of dimensions of fairness for human versus AI competitions in general each representing one category of factors that are often pointed to (in our observations) as the cause of unfairness in the types of human versus AI competitions in games such as the ones analyzed in section~\ref{sec:portrayal}.

   \textbf{Input fairness:} do the two systems have the same input space, e.g.
pixels from a screen? This is especially relevant in electronic games, where it might be the case that a human can only see the limited information on the screen, and needs to make specific actions to gather more information, such as scrolling the viewing window or clicking units to see their attributes. In contrast, an algorithm might have as input a structured list with the location and status of game objects. Even in cases where the input is ostensibly the same (such as an algorithm playing from pixels), is the image capturing apparatus equivalent to a human eye? As an example, the human retina has blind spots and lower peripheral resolution~\cite{strasburger2011peripheral}. Does the computer suffer from the same limitations?

 \textbf{Output fairness:} do both systems have the same output space, that is, the same ways of interacting with the world and with other players? Do they have the same actions
to choose from? The same reaction speed? Does the computer have to
actually press physically buttons or move pieces on a board? Are the systems limited by the same constraints of strength, speed and precision? In cases where communication with other players is possible, do they communicate using the same channels and protocols?

  \textbf{Experience fairness:} have the human and the machine spent the same
amount of time playing the game? Have they played the game under the same
conditions? What about time spent playing closely related games?

    \textbf{Knowledge fairness:} have the agents had access to the same
declarative knowledge (compiled by others) about the game? For example
opening books, pre-trained neural networks, tables of state values?

   \textbf{Compute fairness:} do the agents have the same computational power? How should computational power be measured in this case? A natural metric to use might be power consumption in watts or total energy spent in joules, but it has the shortcoming that, in the case of humans, it is not easy to determine how much power is being used to play the game as opposed to other essential cognitive or bodily functions. The issue gets even more complicated when considering the total energy spent before the match, for example while learning rules and strategies for the game. We could turn to other metrics such as the number of neurons or storage capacity, but it is even harder to establish a meaningful comparison in these cases. Another alternative  could be based on the number or depth of game states explored during tree search, which tends to heavily favor computers.
   
   Finally, the monetary cost of the total computation and the infrastructure required could also be considered, especially in the context of discussing the possibility that a machine might replace humans in a task after achieving \enquote{super-human performance}, which is impractical if deploying and maintaining the artificial agent is several times more expensive than hiring a human to do the same task.
   
   \textbf{Psychological fairness:} are agents subject to the same range of variation in performance due to emotions and states such as fear, joy, mental fatigue, etc.? Currently, this would seem to only apply to humans.

   \textbf{Common-sense fairness:} do the agents have the same knowledge about
other things that may factor into the game? Have they gone to
school? Do they know that dragons are typically dangerous and coins are typically desirable? Have they
seen an American traffic sign? Do they know that cracked walls are more likely to break when you bomb
them? Do they understand instructions given in natural language by NPCs? With current technology, this is a dimension that tends to heavily favor humans.

Note that when it comes to human-versus-human competitions, the dimensions of input and output are usually explicitly addressed as a matter of fairness (such as the creation of divisions and leagues by weight or disability, and the banning of certain performance-altering substances), but variations in the remaining dimensions are usually treated as part of individual player skill. A human player might be praised for having more knowledge of the game, calculating more moves ahead or handling stress better than their opponent, but a machine with advantages in the same areas might be seen as unfair.

% Presumably, in the case of physical abilities, humans are sometimes seen as habing presumably because the machine is seen as having a bigger advantage in this domain than what is possible between humans, whereas the mental and psychological differences between humans are perceived as being in a reasonable range (which does not happen with physical differences).

Each of the dimensions is actually a continuum rather than a binary, and, as will be discussed below, not all of them are equally relevant to all games. Regardless, it should be clear at this point that even determining what constitutes fairness at a game  is a challenging task. We will argue that achieving complete fairness in all dimensions would require building a system that is essentially equivalent to a real, flesh-and-blood human with a full history of human-like experiences.

% The only way to turn up the fairness to 100 on all dimensions is to
% actually create a real, flesh-and-blood human. So no AI/human
% game-playing comparisons will be completely fair on all dimensions.
% QED.

\section{A discussion of fairness in human-AI game benchmarks}
\label{sec:discussion}

At a first glance, the issue of fair conditions in between a computer agent and a human seems more tractable in tabletop games, such as Backgammon, Chess and Go, than in electronic games. A major difference between the two domains seems to be that what we called Input and Output fairness are less relevant in tabletop games. As an example, AlphaGo required a human facilitator to input the current board state into the system, and to apply the move selected by the algorithm to the physical board.  Modifying the system with a camera to read the board state and a robotic arm to move the pieces  might be interesting Computer Vision and Robotic problems on their own, but it would be hard to argue that it would constitute a better Go player or that the competition with humans would be fairer.

Due to this significant difference, we divide discussion below between tabletop games and electronic games. Issues discussed for tabletop games in general also apply to electronic games, but the reverse is not necessarily true. 

Another category that could be considered is that of games involving  direct interaction with the real world, such as the aforementioned Robocup~\cite{PHILOSOC10}, where robots play soccer against robots. These raise even more questions about fairness in the sensorimotor sense, as the physical speed, strength, weight and dimensions of the robots have to be constrained to a similar range than that of humans. However, an in-depth discussion of this topic is outside the scope of this paper and we will focus the remaining discussion on tabletop and electronic games such as the ones described in section~\ref{sec:portrayal}  

\subsection{Fairness in Tabletop games}

The first key issue affecting the fairness between human and artificial players in tabletop games are feelings such as fatigue, fear, anxiety, etc. In~\cite{Swift}, Kasparov comments on the role these factors can play in a match. Ke Jie~\cite{KeJie}, another prominent Go player who has also lost to AlphaGo, stated that psychological factors are possibly \enquote{the weakest part of human beings}. Additionally, sports commentators regularly build a narrative around the mental factors going into an important match, especially one where a lot of pride or money is involved. The magnitude of the psychological effect is unclear from this brief study, but, to whatever degree it might change the outcomes, compensating for it is intractable with current technology. There is no straightforward way to account for these emotions in a computer simulation, and attempting to do so (e.g. by artificially injecting noise in the algorithm's evaluation in situations of high stress) would defeat the purpose of building the best possible game-playing systems. 

A second issue that can be raised is the use of look-up tables for specific points of a match, such as the opening and endgame, and the availability of information about a specific opponent in a match. These could be seen as a matter of Knowledge fairness. Look-up tables have been used in Deep Blue~\cite{campbell2002deep} and suggested as a potential improvement for TD-Gammon's identified weakness in endgame situations.~\cite{tesauro1995temporal}. The use of similar resources in most competitive matches between humans is banned, but when playing versus a computer, should a human have access to the same tables that are available to the algorithm? Similarly, if an algorithm is capable of studying examples of human play in general (as is the case for the original AlphaGo~\cite{silver2016mastering}) or even have some of its parameters or design decisions tuned to face a specific human player (as happened with Deep Blue~\cite{campbell2002deep}), wouldn't it be fair for a human to review a large number of games by an artificial agent, receive a detailed summary of its preferred openings and strategies, perhaps even inspect the source code?

In the same vein, the use of a forward model to simulate future game states, as is done during tree search in Deep Blue~\cite{campbell2002deep} and the rollouts of AlphaGo~\cite{silver2016mastering} could be considered an issue of Compute fairness . This could be compared to giving a human a set of extra boards and pieces with which to simulate potential lines of play during a match, which is also not allowed in competitive play. An important observation is that while it is possible for a human to fully simulate a game of Chess or Go in this way, it is harder to do the same for games that involve randomness and hidden information, and simulation becomes even harder for an unassisted human if the task involves continuous dimensions such as time and distance. 

%We would like to relate these issues to what is called System 1 thinking and System 2 thinking in dual-process theory~\cite{evans1984heuristic}. System 1 thinking, often called intuition or heuristic thinking, has been related in the Reinforcement Learning context to the selection of actions without lookahead~\cite{anthony2017thinking}, such as using a look-up table for openings or a neural network pre-trained on a database of game states. System 2 thinking consists of conscious analytical reasoning, and has been related to Tree Search~\cite{anthony2017thinking}. 

We would like to relate these issues to what is called System 1 thinking and System 2 thinking in dual-process theory~\cite{evans1984heuristic}. System 1 thinking, often called intuition or heuristic thinking, has been related in the Reinforcement Learning context to the selection of actions without lookahead~\cite{anthony2017thinking}, such as using a look-up table for openings or a neural network pre-trained on a database of game states. System 2 thinking consists of conscious analytical reasoning, and has been related to Tree Search~\cite{anthony2017thinking}. While both types of thinking  could be augmented for humans through the use of external tools such as notes on a piece of paper or extra boards for simulation of lines of play, an argument could be made following the Extended Mind~\cite{clark1998extended} that whether such resources are internal or external to a system makes little difference when considering the system's cognitive abilities. Taking this argument to the extreme, we could imagine a situation where a complete artificial game-playing system is viewed as mere augmentation of a human's cognitive abilities, leading  to the absurd scenario of a "human versus AI" match where nonetheless all moves are selected by the same algorithm, one playing for itself, the other in the human's stead. 

In the opposite direction, we could attempt to reduce the computer's advantage by restricting what kinds of techniques it is allowed to use, disallowing the ones that are viewed as inherently unfair. Ultimately, however, by a line of reasoning similar to the Chinese Room thought experiment~\cite{searle1999chinese} , all of the aforementioned advantages could be reduced to following instructions on a piece of paper or doing \textit{mere calculations} and so no AI achievements could ever be considered as proof of \textit{true} mastery in a game. 

A final issue, unrelated to the use of a forward model, is the machine's ability to generalize what it learned across different games or variations of the same game. According to Brooks~\cite{SevenSins}, humans are prone to infer competence from performance. As humans, we might expect a system that performs as the best Go player in the world to be competent enough to play on a board of different dimensions, or play with a different goal (such as the intent to lose) or be at least a passable player in another similar game (such as chess). Marcus~\cite{marcus2018innateness} points out that this is not the case with most existing techniques, and addressing this issue is the motivation behind competition frameworks such as ALE~\cite{bellemare2013arcade}, GVGAI~\cite{perez2016general} and GGP~\cite{genesereth2005general}. While this doesn't strictly affect the fairness of competitions based on playing a single game, with a single ruleset, it is an important point to consider against the narrative that sees the success of AI in a new task as evidence that AGI is just around the corner.

%%% This needs to be rewritten 
 
% The second and third issues could be alleviated as the methods behind existing engines rely less on human examples and pre-calculated lookup tables. Similarly, all three issues can be alleviated as the gap widen between state of the art game-playing agents and human players. In a close series such as the one between Deep Blue and Kasparov in 1997, it is conceivable that fatigue, anxiety, endgame databases or specific opponent knowledge could have played a significant role in the end result. However, today's top chess engines (over 20 years later) play at an estimated ELO rating of 3500~\cite{silver2017mastering2} compared to top humans around 2800~\cite{FIDE}).  However, these concerns about the fairness of comparison between machines and humans could still be relevant for new tasks where human performance might be achieved in the near future, where the machine's margin of victory is potentially still small.

\subsection{Fairness in Electronic Games}

The major difference between tabletop games and electronic games when it comes to perception of fairness seems to be rooted on  the representation of the observation and action space, as well as reaction time, as discussed in ~\cite{Motherboard, Cook}. These are related to what we call Input and Output fairness.

Regarding the observation space, a common paradigm suggestion to achieve greater fairness is playing the game from pixels, rather than from higher level game features. This is the approach followed by Vizdoom~\cite{kempka2016vizdoom} and ALE~\cite{bellemare2013arcade}. While the approach can be said to more closely emulate the way humans perceive video games, the comparison is not perfect.

On one hand, favoring the AI, questions such as \enquote{is the distance between these two objects smaller than the range of my spell?} are still much easier to answer accurately for an agent playing from pixels than for a human. On the other hand, when a human sees pixels in the shape of a coin, a spider and fire, they can reasonably infer that the first object has to be collected, the second attacked and the third avoided, and such heuristic would work well for many games (what we call the common-sense dimension of fairness). Embedding this representation and real-world knowledge in a visual AI system is an unsolved problem, which provides humans with an advantage that is not easy to surmount at the moment.

While objections to high-level representations are valid, taken to the extreme, these objections would imply that no meaningful advancements could be made in video game-playing AI before the field of computer vision is essentially solved. This would be disappointing from a game AI perspective. After all, low-level recognition of pixel patterns is not what immediately comes to mind when we picture a human expertly playing a game. Results obtained on less structured or more general representations can arguably be characterized as more impressive, but the challenges involved in dealing with lower level representations don't necessarily capture what makes games such interesting AI problems in the first place.

For this reason, considerations about the input representation shouldn't be a barrier for game AI research, especially in environments where humans currently have the upper hand. We believe novel results using higher level representations are important, and further research that attempts to replicate these results while using less favorable or more general representations are also important and will likely naturally follow the initial results.

Similarly, the representation of the action space can take many forms, such as high-level representations like used in OpenAI's methodology~\cite{OpenAI5} where an action is viewed as a tuple  (e.g. [ability, target, offset]), a simulation of user interface commands such as screen movements and unit highlighting in Starcraft~\cite{vinyals2017starcraft} and the direct simulation of aa virtual controller as in ALE~\cite{bellemare2013arcade}. The extreme position would be to insist on a robotic arm manipulating a physical controller or keyboard, which would again distract researchers from other legitimate game AI problems that can be tackled with higher level representations.

Excessively fast reaction speed is often cited as one of the factors that make an agent play in a perceived artificial fashion~\cite{khalifa2016modifying}.  A popular solution, used by OpenAI~\cite{OpenAI5} is to directly enforce a specific reaction time. Alternate solutions involve the "Sticky Action" and other methods discussed in~\cite{machado2017revisiting} for the ALE environment. Interestingly, the  original motivation of Sticky Actions in that paper is not to emulate human play, but to provide enough randomness to the otherwise deterministic ALE environment. This forces the agent to learn "closed loop policies" that react to a perceived game state, rather than potential "open loop policies" that merely memorize effective action sequences. Regardless, this also works to avoid inhuman reaction speeds.

\section{Conclusion}

We have briefly recapitulated some of the most important game AI benchmark results in the past three decades, for both tabletop games (Backgammon, Chess and Go) and electronic games (specially Starcraft and Dota 2) and looked at some of the claims made by the authors of these game-playing systems and third-party comments made by general media and research communities. From those, we conclude that there is a tendency to extrapolate from AI achievements in game Benchmarks to speculation about Artificial General Intelligence (AGI) scenarios where AI will eventually beat humans in all or most tasks. We have also seen examples of public concerns about the fairness of these benchmarks.

We have proposed a taxonomy of dimensions on which to evaluate fairness in a competition between two game-playing systems (such as a human and an AI agent).  We provided examples of how these apply to tabletop games and electronic games, noting that there is greater focus on the Input and Output dimensions of fairness for electronic games.

Ultimately, we argue that there are so many possible games, and so many possible architectures of game-playing agents, differing so widely in the dimensions of fairness, that it is impossible to infer human-level intelligence from success in any single game, and that a completely fair competition can only be achieved against an artificial system that is essentially equivalent  to a flesh and blood human.

This conclusion should not serve to infer that, if complete fairness is impossible, accomplishments in game-playing AI are meaningless in general, or that benchmarks that feature many differences between humans and AI along our fairness dimensions have no value. We observe that, usually, when a significant benchmark is reached, research often follows in order to make the system less reliant on human expertise, more sample-efficient, less reliant on  extremely fast reaction speeds and more generalizable to similar problems, leading to systems with fewer restrictions and wider applications. 

It is also important to highlight that research in AI for games does not happen in a vacuum, and is often used as a stepping stone to solve complex real-world problems. While the ability of an agent to make a large number of actions per minute or to access a vast amount of information can seem unfair in a game context, this could be extremely desirable when applied to real-world problems that require a high frequency of actions and with access to more information than could conceivably be processed by humans. Finally, we acknowledge that discussions of fairness can have complex political implications, especially when it comes to systems with so-called \enquote{super-human} abilities being employed in the real world. While these considerations are out of the scope of this paper, we believe this is an important discussion to be had in the research community.

\section{Acknowledgements}
RC gratefully acknowledges the financial support from Honda Research Institute Europe (HRI-EU). CS is funded by the EU Horizon 2020 programme / Marie Sklodowska- Curie grant 705643. 

% We also listed and discussed some factors that might affect how fairness between human and AI. For both tabletop and electronic games, human feelings such as fatigue, fear and anxiety, the ability of artificial to study the style of a specific human opponent and techniques such as databases of moves can be construed as unfair advantages of AI. However, as the gap between the best agents and humans widens (for games where the AI has an advantage) , it becomes harder to argue that these are the deciding factors in the outcome.

% Specifically to video games, most of the discussion is centered around the interface used by the agent for input and output, and also on the unfairness of a reaction time that is quick beyond human. We argue that (all else being equal) results using architectures that interact with the game in a human-like fashion are more impressive, this should not be a discouragement to research done using more high-level representations. It is likely that, once these games are eventually beaten using more generous architectures, work that attempts to achieve similar performance while reducing the AI program's inherent advantages will quickly follow.